%% file: main.tex
\input{styles}
\title{Principled Koopman Representations with Kalman Inference for Efficient Time-Series Prediction}

\author {
    Ruiquan Li\textsuperscript{\rm 1},
    Yuheng Bu\textsuperscript{\rm 1}
}
\affiliations {
    \textsuperscript{\rm 1}Department of Computer Science, UC Santa Barbara\\
    rli667@ucsb.edu, buyuheng@ucsb.edu
}

\begin{document}
\maketitle


\begin{abstract}
The Koopman operator has been widely used for time-series prediction in dynamical systems. However, prior work that learns latent ``Koopman spaces'' using neural networks often did not construct a valid Koopman space for forecasting, as these representations may be mathematically inconsistent with the operator-theoretic formulation and fail to capture the intrinsic low-rank structure of system dynamics. To address this issue, we introduce K$^2$SVD, a method that explicitly learns the leading singular functions of the Koopman operator by optimizing a Hilbert-Schmidt objective. This yields a well-defined low-rank approximation of the Koopman operator with an interpretable linear combination, featuring a compact latent space with less than $10\%$ of the dimensions used in previous work. In the learned Koopman space, K$^2$SVD further captures temporal evolution with a linear Gaussian state-space model and performs inference via Kalman filtering, mitigating noise accumulation during multi-step prediction. Empirical results show that K$^2$SVD outperforms state-of-the-art methods across multiple datasets, with significantly faster prediction speeds and lower computational cost than previous efficiency-focused models. This highlights the benefits of principled low-rank Koopman representations and opens up broader potential for applications.


\end{abstract}

\section{Introduction}

Time-series forecasting is a fundamental problem across a wide range of scientific and industrial domains, often requiring models to handle highly nonlinear transformations under noisy perturbations. Recent deep learning models, including transformers~\cite{li2019enhancing}, sequence-to-sequence architectures, and foundation models~\cite{das2024decoder,ansari2024chronos}, have achieved strong empirical performance. However, in domains such as economics, weather, and energy, where explainable and expressive latent spaces are crucial, purely black-box prediction models are often insufficient.


Koopman operator \cite{lan2013linearization} offers a compelling alternative to black-box models by providing a mathematically grounded and expressive latent space. By lifting data into an infinite-dimensional space of measurement functions, the Koopman framework represents nonlinear dynamics through a linear operator that maps measurements to their next-step evolution. This perspective also enables spectral methods to analyze the widely observed low-rank structure in system dynamics, bridging classical dynamical systems theory and modern machine learning. In particular, recent Koopman singular value decomposition (SVD)-based methods \cite{jeong2025efficient,kostic2024learning,wu2020variational} learn the leading singular functions of the operator, demonstrating the effectiveness of Koopman representations for modeling dynamical systems.

Despite these advances, many existing approaches to probabilistic time-series prediction \cite{liu2026koopman_kalman,liu2023koopa} that claim to construct a Koopman space rely on heuristic or loosely defined latent representations, and do not rigorously adhere to the underlying operator-theoretic framework. In particular, the learned embeddings are often optimized for reconstruction loss solely rather than for satisfying the definition of Koopman-invariant subspaces. As a result, they may fail to preserve linear evolution in a well-defined function space and may require redundant modules to handle nonlinearity. Consequently, the resulting ``Koopman spaces'' may not admit a consistent operator interpretation, and their associated dynamics can deviate from the true spectral structure, especially the low-rank structure, leading to parameter redundancy and inefficient prediction.


\begin{figure}[h]   
    \centering
    \includegraphics[width=0.45\textwidth]{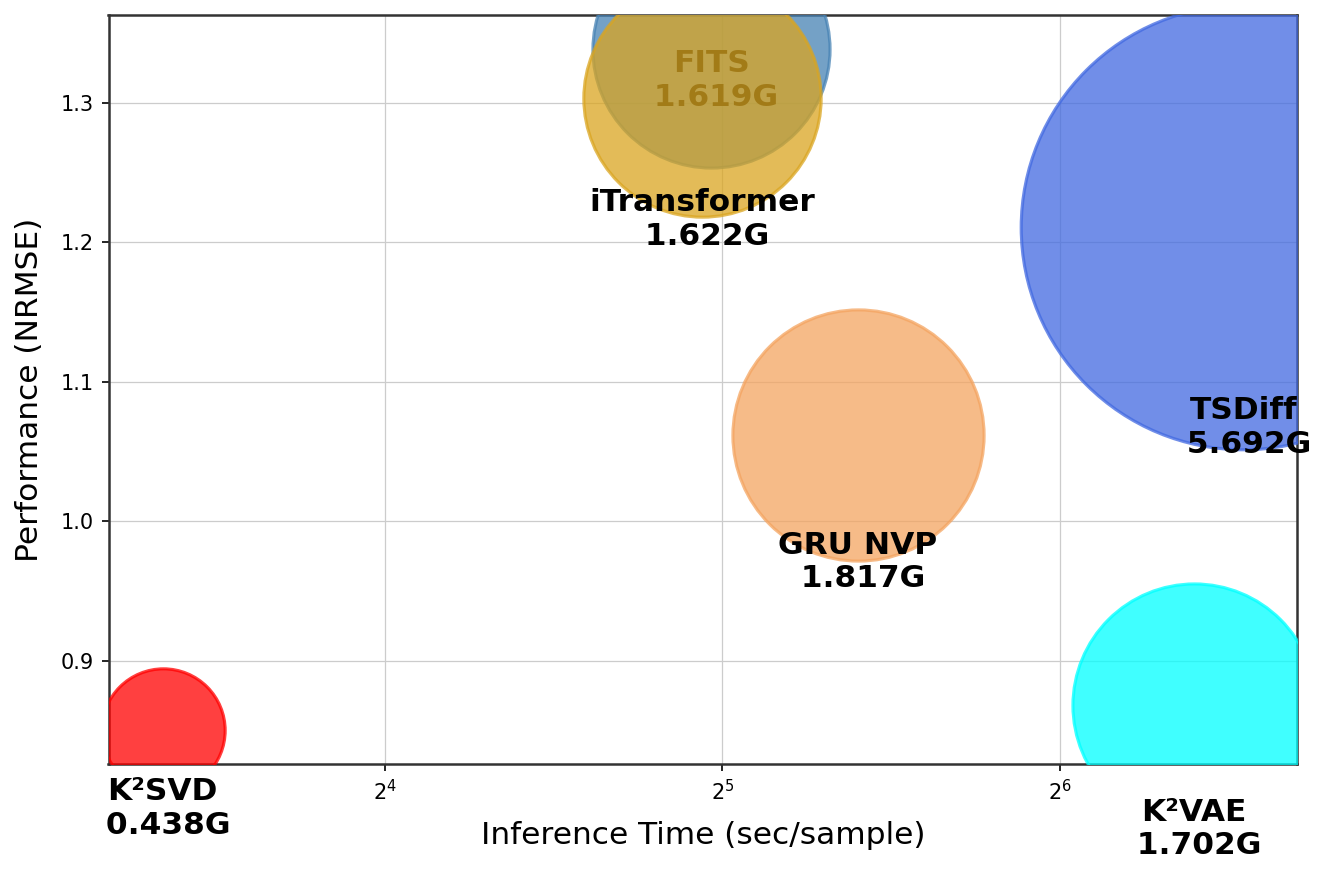}
    \caption{Comparison of predictive performance, GPU usage, and inference time on the Electricity dataset. K$^2$SVD outperforms competing methods in time and space efficiency, as well as overall performance. Circle size indicates peak allocated GPU memory during training.}
    \label{fig:first_img}
    \vspace{-1em}
\end{figure}
To address these limitations, we propose K$^2$SVD, a principled framework that explicitly enforces a low-rank Koopman structure by learning approximations to its leading singular functions. Rather than treating the latent space as an unconstrained embedding, our method constructs a theoretically grounded representation by optimizing a Hilbert-Schmidt objective that directly targets the best rank-$k$ approximation of the operator. This yields a latent space with well-defined linear dynamics, enabling faithful spectral characterization and linear temporal evolution, which also reduces the computation dramatically.
In the Koopman latent space, temporal evolution is modeled as a linear transition with nonlinear perturbations. Given a principled Koopman space, we posit that these nonlinear residuals have low magnitude and limited predictive information, and can therefore be treated as Gaussian noise to be filtered. Furthermore, due to the iterative nature of time-series prediction, we perform inference via Kalman filtering to mitigate noise accumulation during the forward pass.

The overall model is trained using a two-stage scheme that first establishes the Koopman subspace and then refines the dynamical and observation components using Kalman filter in an end-to-end manner. This design combines theoretical rigor with practical scalability, yielding a model that captures low-rank dynamics while maintaining strong forecasting performance and parameter efficiency. As shown in Figure~\ref{fig:first_img}, K$^2$SVD achieves competitive prediction performance with substantially low GPU memory usage and shorter inference time than existing methods.
Our contribution is as follows:

\begin{itemize}[itemsep=0pt,topsep=-2pt, leftmargin=10pt]
    \item We propose K$^2$SVD, a principled Koopman forecasting framework that learns a low-rank Koopman space by approximating the leading singular functions of the Koopman operator through a Hilbert-Schmidt objective, yielding a mathematically grounded latent space.

    \item We integrate the learned Koopman representation with a linear Gaussian state-space model and Kalman filtering, enabling efficient multi-step inference that mitigates noise accumulation.

    \item We empirically show that K$^2$SVD achieves competitive forecasting performance across multiple dynamic systems and real-world time-series benchmarks while requiring substantially fewer parameters and shorter inference time than existing methods, demonstrating the benefit of explicitly modeling low-rank Koopman structure.
\end{itemize}

\section{Related Works}

\textbf{Probabilistic time series forecasting.} Time series prediction has evolved from classical autoregressive models to deep generative frameworks. Early neural approaches such as DeepAR~\cite{salinas2020deepar}, N-BEATS~\cite{oreshkin2020nbeats}, and transformer-based models learn flexible sequence representations for multi-horizon forecasting~\cite{lim2021tft,zhou2021informer,rasul2021multivariate}. More recent works incorporate structured dynamical priors to improve long-term prediction and interpretability.

\textbf{Koopman operator learning.}
The Koopman operator framework provides a principled approach to analyzing nonlinear dynamical systems by lifting them into a linear space. Classical works establish their spectral foundations and connections to ergodic theory and modal decomposition~\cite{mezic2005koopman,mezic2013analysis,rowley2009spectral}. Building on this theory, data-driven approximations such as dynamic mode decomposition (DMD) and its extensions enable finite-dimensional representations of the Koopman operator directly from data~\cite{schmid2010dmd,tu2013dynamic,williams2015data, williams2015kernel}. In particular, low-rank approximations based on SVD play a central role in extracting dominant Koopman modes, and more recently, learning-based approaches leverage neural networks to recover Koopman subspaces~\cite{wu2020variational,mardt2018vampnets,kostic2024learning}. These methods highlight the importance of low-rank Koopman representations and SVD approximations for high-dimensional dynamical systems.

\textbf{Koopman-Based Probabilistic Forecasting.} Koopman-based models have gained increasing attention. Koopman Neural Forecaster (KNF) integrates Koopman embeddings with distributional forecasting to handle temporal distribution shifts~\cite{wang2022knf}, while Koopa introduces hierarchical Koopman predictors to decompose non-stationary dynamics into invariant and time-varying components~\cite{liu2023koopa}. Related generative approaches, such as Koopman VAEs, embed linear dynamics into latent representations for stable sequence generation~\cite{naiman2023koopmanvae}. Hybrid probabilistic models further combine Koopman theory with latent-variable methods, including Kalman filtering. These methods typically use Koopman lifting with Kalman state-space updates: K-ESKF~\cite{huang2024koopman} improves state estimation during highly agile motion, while K$^2$VAE~\cite{wu2025k2vaekoopmankalmanenhancedvariational} uses variational autoencoder training for long-term prediction. Together, these works show that combining Koopman theory with probabilistic deep learning offers a powerful framework for modeling complex temporal dynamics under uncertainty.

\section{Preliminaries}
\textbf{Koopman theory.} Consider a stochastic discrete-time dynamical system
\begin{equation}
    \mathbf{x}_{t+1} = \xi(F(\mathbf{x}_t), \epsilon_t).\label{eq:obj}
\end{equation}
Here, $F: \mathcal{X} \rightarrow \mathcal{X}$ is a possibly nonlinear mapping for a domain $\mathcal{X} \subseteq \mathbb{R}^d$, $\epsilon_t \sim \mathcal{D}$ denotes independent random noise.
Assuming that $F$ and the noise distribution $\mathcal{D}$ are time-invariant, the process becomes a time-homogeneous Markov process with transition density $p(\mathbf{x}' \mid \mathbf{x})$, i.e., $\Pr(\mathbf{X}_{t+1} \in A \mid \mathbf{X}_t = \mathbf{x}) = \int_A p(\mathbf{x}' \mid \mathbf{x})\, d\mathbf{x}'$ for all measurable sets $A \subseteq \mathcal{X}$ and all $t \geq 0$.

In the stochastic setup, the dynamics is fully captured by the transition density $p(\mathbf{x}' \mid \mathbf{x})$, which is induced by $(\xi \circ F)(\cdot)$, and thus the problem becomes analyzing $p(\mathbf{x}' \mid \mathbf{x})$ of a Markov chain. Here, the Koopman operator becomes the conditional expectation operator, i.e., for an observable $g: \mathcal{X} \rightarrow \mathbb{R}$,
\begin{equation}
    (\mathcal{K}g)(\mathbf{x}) \triangleq \mathbb{E}_{p(\mathbf{x}' \mid \mathbf{x})}[g(\mathbf{x}')].
\end{equation}
By the Markov property, repeatedly applying the Koopman operator will correspond to the \textit{multi-step prediction} in terms of the posterior mean of $g(\mathbf{X}_t)$ given $\mathbf{X}_0 = \mathbf{x}_0$, i.e., $(\mathcal{K}^t g)(\mathbf{x}_0) = \mathbb{E}_{p(\mathbf{x}_t \mid \mathbf{x}_0)}[g(\mathbf{x}_t)]$. We will assume that the operator is compact throughout this paper.

\textbf{Problem setting. }Given historical data $X = (\mathbf{x}_1, \mathbf{x}_2, \cdots, \mathbf{x}_T)\in \mathbb{R}^{N\times T}$, our goal is to model the stochastic process and predict the future sequence $Y = (\mathbf{x}_1, \mathbf{x}_2, \cdots,\mathbf{x}_{T+1}, \mathbf{x}_{T+2}, \cdots, \mathbf{x}_{T+t})$ sampling each data from their conditional distribution $p(\mathbf{x}_t| \mathbf{x}_0)$. Suppose two data points at different time spots $\mathbf{x}, \mathbf{x}'$, there marginal distributions are $\rho_0, \rho_1$ satisfies $\rho_1(\mathbf{x}') \triangleq \mathbb{E}_{\rho_0(\mathbf{x})}[p(\mathbf{x}'|\mathbf{x})]$.

Let $\mathcal{X}$ be a measurable space and let $\rho_0$ and $\rho_1$ be (finite) measures on $\mathcal{X}$ representing the distributions of the current and future states, respectively. For any measure $\rho$ on $\mathcal{X}$ define $L^2_\rho(\mathcal{X}) \triangleq \{f: \mathcal{X} \rightarrow \mathbb{C} \mid \int |f(\mathbf{x})|^2 \rho(d\mathbf{x}) < \infty\}$. Equipped with the inner product $\langle f, g \rangle_\rho \triangleq \int f(\mathbf{x})g(\mathbf{x})\rho(d\mathbf{x})$, $L^2_\rho(\mathcal{X})$ is a Hilbert space. The Koopman operator $\mathcal{K}$ is then a mapping from $L^2_{\rho_1}(\mathcal{X})$ to $L^2_{\rho_0}(\mathcal{X})$.

\textbf{SVD of Operators.} Given $\mathcal{K}: L^2_{\rho_1}(\mathcal{X}) \rightarrow L^2_{\rho_0}(\mathcal{X})$, the Singular Value decomposition(SVD) decomposes $\mathcal{K}$ into a countable sum of rank-1 operators:
\begin{equation}
    \mathcal{K} = \sum_{i=1}^{\infty} \sigma_i \, f_i \otimes g_i,
\end{equation}
where the outer product operator acts as $(f_i \otimes g_i)h = \langle g_i, h \rangle_{\rho_1} f_i$. The singular values and singular functions satisfy:
\begin{equation}
    \mathcal{K} g_i = \sigma_i f_i, \qquad \mathcal{K}^* f_i = \sigma_i g_i,
\end{equation}
with orthogonality conditions:
\begin{equation}
    \langle g_i, g_j \rangle_{\rho_1} = \delta_{ij}, \qquad \langle f_i, f_j \rangle_{\rho_0} = \delta_{ij},
\end{equation}
where $\{g_i\} \subset L^2_{\rho_1}(\mathcal{X})$ forms an orthonormal basis of the input space (observables at current state $\mathbf{x}_t$), and $\{f_i\} \subset L^2_{\rho_0}(\mathcal{X})$ forms an orthonormal basis of the output space (observables at future state $\mathbf{x}_{t+1}$), with singular values $\sigma_1 \geq \sigma_2 \geq \cdots \geq 0$. In a real-world scenario, the transition often processes \textbf{Low-Rank (LoRA) properties}, which implies $\sigma_i$ will decline to $0$ rapidly as $i$ increases.

\textit{Second-moment matrices} are key quantities in SVD. We thus introduce the following shorthand notation for convenience. The \textit{second-moment matrix} of vector-valued functions $\mathbf{h}_1(\cdot)$ and $\mathbf{h}_2(\cdot)$ with respect to a distribution $\rho$ is denoted as
\begin{equation}
    \mathsf{M}_{\rho}[\mathbf{h}_1, \mathbf{h}_2] \triangleq \mathbb{E}_{\rho(\mathbf{x})}[\mathbf{h}_1(\mathbf{x})\mathbf{h}_2(\mathbf{x})^\top],
\end{equation}
and we write $\mathsf{M}_{\rho}[\mathbf{h}] \triangleq \mathsf{M}_{\rho}[\mathbf{h}, \mathbf{h}]$ for shorthand. The \textit{joint second-moment matrix} of $\mathbf{h}_1(\cdot)$ and $\mathbf{h}_2(\cdot)$ over the joint distribution $\rho_0(\mathbf{x})p(\mathbf{x}' \mid \mathbf{x})$ is defined and denoted as
\begin{equation}
    \mathsf{T}[\mathbf{h}_1, \mathbf{h}_2] \triangleq \mathbb{E}_{\rho_0(\mathbf{x})p(\mathbf{x}' \mid \mathbf{x})}[\mathbf{h}_1(\mathbf{x})\mathbf{h}_2(\mathbf{x}')^\top],
\end{equation}
which satisfies the identities $\mathsf{T}[\mathbf{h}_1, \mathbf{h}_2] = \mathsf{M}_{\rho_0}[\mathbf{h}_1, \mathcal{K}\mathbf{h}_2] = \mathsf{M}_{\rho_1}[\mathcal{K}^*\mathbf{h}_1, \mathbf{h}_2]$.
In practice, these quantities are to be estimated with trajectories, and their empirical estimates are denoted with hats ($\hat{\phantom{x}}$), i.e., $\hat{\mathsf{M}}_{\rho_0}[\mathbf{f}]$, $\hat{\mathsf{M}}_{\rho_1}[\mathbf{g}]$, and $\hat{\mathsf{T}}[\mathbf{f}, \mathbf{g}]$.

\section{Proposed Method}

\begin{figure*}[t]
    \centering
    \includegraphics[width=0.9\textwidth]{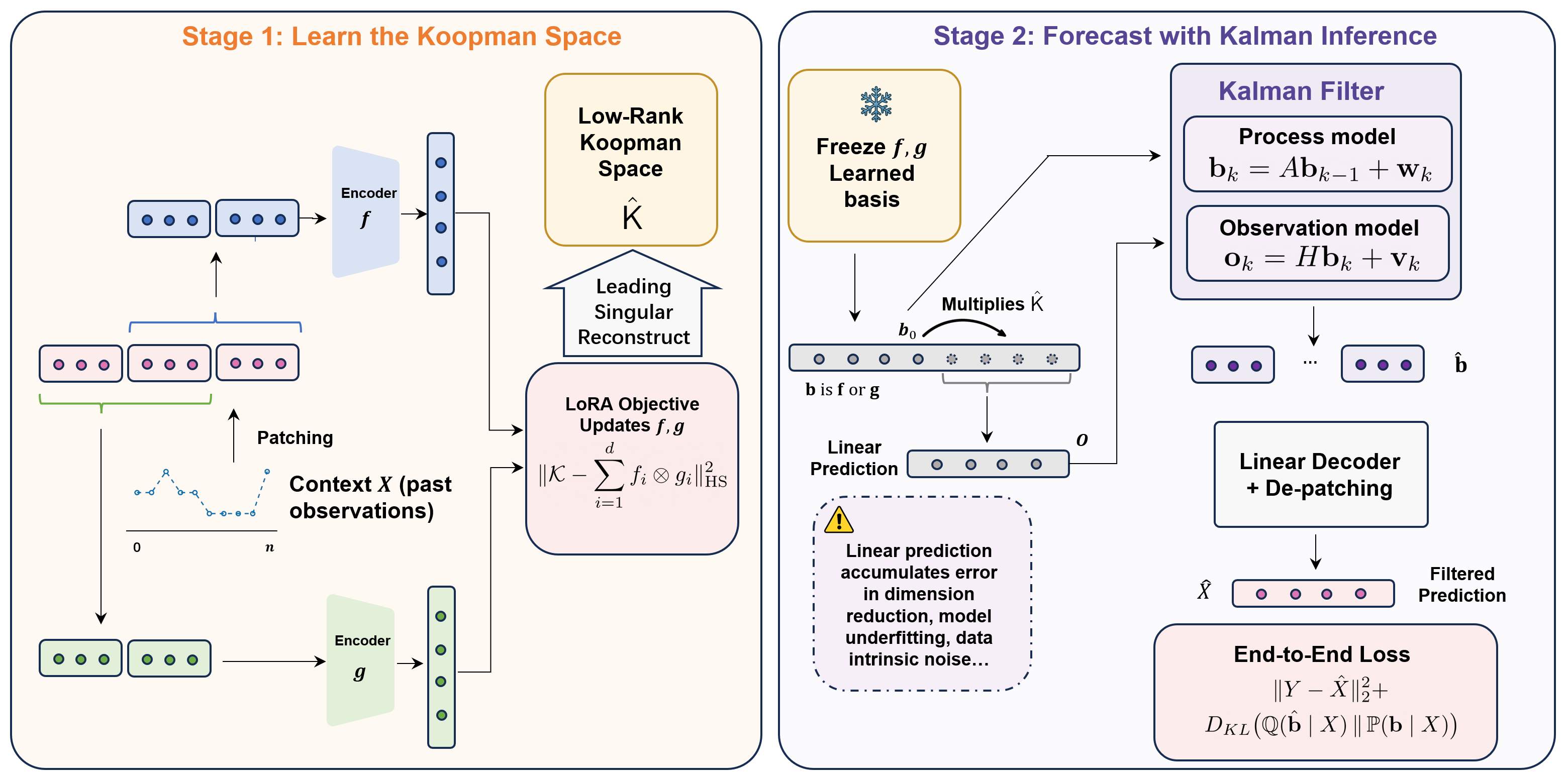}
    \caption{Overview of the two-stage K$^2$SVD procedure for learning a low-rank Koopman space and performing Kalman inference.}
    \vspace{-1em}
    \label{fig:model}
\end{figure*}

We propose K$^2$SVD, a time-series forecasting framework that combines Koopman representations with Kalman inference to model complex nonlinear dynamics through linear evolution in a latent space. Specifically, K$^2$SVD learns a neural representation that lifts observations into a low-rank Koopman space, where state transitions are modeled as linear dynamics perturbed by Gaussian noise. This enables efficient and principled inference via the Kalman filter.

K$^2$SVD is trained with a two-stage algorithm: first, it learns the low-rank Koopman space by optimizing the Hilbert--Schmidt objective; second, it freezes the learned Koopman basis and trains the Kalman inference and observation components for forecasting. The predicted latent states are then mapped back to the observation space through a linear projection. This design combines neural expressivity with the analytical tractability of linear dynamical systems. The full procedure is illustrated in Figure~\ref{fig:model} and the pipeline is summarized in Appendix~\ref{ref:algo}.

\subsection{First Stage: Learning the Koopman Space}
\subsubsection{Encoder}

We consider \textbf{multivariate patches} as tokens to implicitly model the cross-variable interaction during state transition. We divide the context series
\[
X = [\mathbf{x}_1, \mathbf{x}_2, \cdots, \mathbf{x}_T] \in \mathbb{R}^{N \times T}
\]
into non-overlapping patches:
\[
X^P = [\mathbf{x}_1^P, \mathbf{x}_2^P, \cdots, \mathbf{x}_n^P] \in \mathbb{R}^{N \times s \times n},
\]
where $s = T/n$ denotes the patch size, $n$ is the number of patches, and patch $\mathbf{x}_i^P \in \mathbb{R}^{N \times s}$. Similarly, we patchwise the horizon as well with the same patch size $s$ as follows:
\[
Y^P = [\mathbf{x}_1^P, \mathbf{x}_2^P, \cdots, \mathbf{x}_m^P] \in \mathbb{R}^{N \times s \times m},
\]
In the following, we capture the singular functions of the desired Koopman space with neural networks: $\mathbf{f}(X^P) = [\mathbf{f} (\mathbf{x}_1^P), \mathbf{f}(\mathbf{x}_2^P),\cdots, \mathbf{f}(\mathbf{x}_n^P)]$, and for similar cases $\mathbf{g}(X^P) = [\mathbf{g} (\mathbf{x}_1^P), \mathbf{g}(\mathbf{x}_2^P),\cdots, \mathbf{g}(\mathbf{x}_n^P)]$ where $\mathbf{f}, \mathbf{g}$ are desired neural network corresponding to the intrinsic property of data.

\subsubsection{LoRA Loss Function}
The best rank-$d$ approximation of $\mathcal{K}$ in the Hilbert-Schmidt norm is:
\begin{equation}
    \mathcal{K} \approx \sum_{i=1}^{d} \sigma_i \, f_i(\mathbf{x}) \otimes g_i(\mathbf{x}')
\end{equation}
with the learned $f_i$ and $g_i$ spans the optimal approximated rank $d$ subspace $\mathcal{H}^d$ 
of the output and input space.
To bypass the numerical instability and optimization challenges, \cite{jeong2025efficient} propose to directly minimize the low-rank approximation error $\|\mathcal{K} - \sum_{i=1}^{d} f_i \otimes g_i\|^2_{\text{HS}}$ to find the singular functions of the low-rank Koopman operator Hilbert space, where $\| \cdot \|_{HS}$ denotes the Hilbert–Schmidt loss. Succinctly, the learning objective can be expressed as:
\begin{equation}
    \mathcal{L}_{\text{lora}}(\tilde{\mathbf{f}}, \tilde{\mathbf{g}}) \triangleq -2\,\text{tr}(\mathsf{T}[\tilde{\mathbf{f}}, \tilde{\mathbf{g}}]) + \text{tr}(\mathsf{M}_{\rho_0}[\tilde{\mathbf{f}}]\mathsf{M}_{\rho_1}[\tilde{\mathbf{g}}]),
\end{equation}
where $ \tilde{\mathbf{f}}\!=\![ \mathbf{f}(\mathbf{x}_2^P), \mathbf{f}(\mathbf{x}_3^P),\cdots, \mathbf{f}(\mathbf{x}_n^P)]$, and $ \tilde{\mathbf{g}}\!=\![ \mathbf{g}(\mathbf{x}_1^P), \mathbf{g}(\mathbf{x}_2^P),\cdots, \mathbf{g}(\mathbf{x}_{n-1}^P)]$. A detailed derivation of the loss is provided in Appendix~\ref{sec:proof}. Noting that the LoRA loss does not require ground-truth prediction for calculating the gradient, we can train it separately in a stage before inference to stabilize and well-structure the Koopman space.

For generality we denote the base of singular space as $\mathbf{b}\in \{\mathbf{f}, \mathbf{g}\}$, and $ \mathbf{b}^P = [\mathbf{b}_1, \mathbf{b}_2, \cdots, \mathbf{b}_n]$. Since the derived singular functions span the best approximated rank $d$ subspace of the input and output space $L^2_{\rho}(\mathcal{X})$, the linear combination of them $\mathbf{h}(X^P)=\sum_{i=0}^{d}z_i \mathbf{b}_i(\mathbf{x}_i^P)$ is also an approximation in $\mathcal{H}^d$ of its corresponding infinite dimensional counterpart. Therefore, there exists a $\mathsf{K}_b : \mathcal{H}^d \to \mathcal{H}^d$ such that
\begin{equation}
    (\mathcal{K}\mathbf{b})(\mathbf{x}) = \mathbb{E}_{p(\mathbf{x}' \mid \mathbf{x})}[\mathbf{b}(\mathbf{x}')] \approx \mathsf{K}_\mathbf{b}\mathbf{b}(\mathbf{x}).
\end{equation}

Ideally, if the data obeys the dynamics given by \eqref{eq:obj}, i.e., has no temporal distribution shift, Markovian, and the latent dimension is sufficient for linearization, the prediction can be performed by a simple linear rollout. Given the linear relation, $\mathsf{K}_b$ can be approximately solved with linear regression $\mathsf{K}^{\text{ols}}_{\mathbf{b}} \triangleq (\mathsf{M}_{\rho_0}[\mathbf{b}])^+\mathsf{T}[\mathbf{b}] \in \mathbb{R}^{k \times k}$ and we multiply it repeatedly to the last state in the training data until the prediction fills $m=(t+T)/s$ time steps to get the linearly reconstructed base vectors of horizon data,
\begin{equation}\label{equ:linear}
    \mathbf{b}^{L}\! = \![\mathsf{K}_\mathbf{b}^{ols}\mathbf{b}(\mathbf{x}^P_n), (\!\mathsf{K}_\mathbf{b}^{ols}\!)^2\mathbf{b}(\mathbf{x}^P_n), \cdots, (\!\mathsf{K}_\mathbf{b}^{ols}\!)^{m\!-\!n}\mathbf{b}(\mathbf{x}^P_n)].
\end{equation}
Noted that $\mathbf{b}^{L}$ has the same length $(m-n)$ as the prediction, and we treat it as a preliminary estimate. As shown in the ablation study in Section~\ref{sec:abl_kalman}, it is not sufficiently accurate, motivating the use of a Kalman filter.

\subsection{Second Stage: Forecast with Kalman Inference}\label{sec:Kalman}

\subsubsection{Kalman Filter for State Space Evolution}
As described above, in real-world datasets, nonlinear residuals can cause misalignment between the actual base vector in the state space and its linear expansion $\mathbf{b}^L$. This noise can arise from dimension reduction, nonlinear or non-Markovian dynamics, random perturbations in the data, and errors in the linear regression estimate $\mathsf{K}_b^{\mathrm{ols}}$. Despite these nonlinear residuals, we posit that the learned Koopman space is sufficiently representative so that the residual components carry limited predictive information and can be mitigated as noise during inference. Considering the recursive nature of time-series prediction, the Kalman filter provides a lightweight, closed-form inference procedure that minimizes estimation error under linear Gaussian dynamics.

For $k\ge n+1$, its processing model is:
\begin{equation}
    \mathbf{b}_k = A\mathbf{b}_{k-1} + \mathbf{w}_k,
\end{equation}
where $A\in d\times d$ is the transition matrix, which we initiate with $\mathsf{K}_b^{ols}$ by definition. $\mathbf{w}_k\sim N(\mathbf{0}, Q)$ is the process noise and $Q$ is its covariance matrix. Accordingly, the observation model is:
\begin{equation}
    \mathbf{o}_k = H\mathbf{b}_{k} + \mathbf{v}_k.
\end{equation}
Here $H\in d\times d$ is the observation matrix, which we initiate with $I$. $\mathbf{o}_k$ is the observation and we use the expanded linear part $\mathbf{b}^{L}[k]$ in~\eqref{equ:linear}, and $\mathbf{v}_k\sim N(\mathbf{0}, R)$.
Then we conduct the prediction Step and update Step iteratively. The prediction step can be formulated as:
   \begin{align}
    \hat{\mathbf{b}}_{k|k-1} &= A\hat{\mathbf{b}}_{k-1|k-1},  \\
    \Sigma_{k|k-1} &= A\Sigma_{k-1|k-1}A^T + Q,
\end{align}
where $\hat{\mathbf{b}}_{k|k-1}$ is the predicted state and $\Sigma_{k|k-1}$ is the predicted covariance matrix of the process uncertainty. The update step then uses the Kalman gain $G_k$ to balance the prediction and observation, thereby refining the latent state:
\begin{align}
    G_k &= \Sigma_{k|k-1} H^T (H\Sigma_{k|k-1} H^T + R)^{-1}, \\
    \hat{\mathbf{b}}_{k|k} &= \hat{\mathbf{b}}_{k|k-1} + G_k(\mathbf{o}_k - H\hat{\mathbf{b}}_{k|k-1}), \label{eq:update}\\
    \Sigma_{k|k} &= (I - G_k H)\Sigma_{k|k-1}.
\end{align}

We note that $A, H, Q, R$ are trainable parameters in the above process. Iteratively, we obtain the refined predicted singular vectors
$\hat{\mathbf{b}} = [\hat{\mathbf{b}}_1, \hat{\mathbf{b}}_2,\cdots,\hat{\mathbf{b}}_m]$.

\subsubsection{Decoder}
Given the learned base functions $\mathbf{b}$ and an observable $h$, we compute the projection coefficient $\mathbf{z}^{ols}_{\mathbf{b}} \triangleq (\mathsf{M}_{\rho_0}[\mathbf{b}])^\dagger \langle h, \mathbf{b}\rangle_{\rho_0} \in \mathbb{R}^{k \times N\times s}$ by solving the linear regression $h(\mathbf{x}) \approx \mathbf{z}^\top \mathbf{b}(\mathbf{x})$. This coefficient is fixed throughout the forecasting process. In our case, we take $h(\mathbf{x})=\mathbf{x}$, so that the predicted data can be decoded from the learned base functions as follows:
\begin{equation}
    \hat{{X}}^P = \hat{\mathbf{b}}^{\top}\mathbf{z}^{ols}_{\mathbf{b}}.
\end{equation}
After prediction, we de-patch the outputs to the prediction $\hat{X}$ and optimize an MSE-based reconstruction loss, together with an additional KL-divergence term on the filter output distribution $\hat{\mathbf{b}}$. This regularization encourages a well-structured latent representation and helps preserve the orthogonality of the learned base functions, leading to the following loss:
\begin{equation}
\mathcal{L}_{e} = \|Y - \hat{X}\|_2^2
+ D_{KL}\big(\mathbb{Q}(\hat{\mathbf{b}}\mid X)\,\|\,\mathbb{P}(\mathbf{b}\mid X)\big),
\end{equation}
where $\mathbb{P}(\mathbf{b}\mid X)\big)=\mathcal{N}(0, \mathbf{I})$, and $\hat{X}$, $\hat{\mathbf{b}}$ are induced by the trainable Kalman filter.

\subsection{Overall Algorithm}
We adopt a two-stage training procedure for learning the Koopman representation. 

\textbf{Train encoders with LoRA objective.} First, we independently train the Koopman module using the LoRA objective to capture the low-rank dynamics and establish a stable latent space for linear evolution. We then freeze the learned functions, assuming that the low-rank structure has been adequately captured, and estimate the Koopman matrix. 

\textbf{Dynamically updated matrix.}
We estimate the Koopman operator $\mathsf{K}_{\mathbf{b}}$ via linear regression and initialize the transition matrix $A$ of the Kalman filter using $\mathsf{K}_\mathbf{b}^{ols}$, averaged over 128 training batches. Because the patched training data provide too few samples for reliable regression,  we instead treat the finite-dimensional Koopman matrix $\mathsf{K}_{\theta}$  as trainable and optimize it jointly in an end-to-end manner using the forecasting objective $\mathcal{L}_e$. The Koopman matrix is initialized using the static least-squares estimate $\mathsf{K}_{\theta} \leftarrow \mathsf{K}_{\mathbf{b}}^{\mathrm{ols}}$.
The proposed K$^2$SVD preserves the principled operator-consistent latent space learned by LoRA, while allowing the transition dynamics to adapt to the downstream multi-step forecasting objective, mitigating the finite-sample
noise engendered in the regression. Since $\mathsf{K}_{\theta}\in\mathbb{R}^{d\times d}$ is low-dimensional, making it trainable adds only a small computational and parameter overhead.


\section{Experiments}
In this section, we present empirical results that validate the effectiveness and performance of K$^2$SVD. We first examine the singular-value spectra of the learned Koopman spaces to characterize the intrinsic rank structure of the datasets. We evaluate forecasting on both low- and high-dimensional datasets, with an emphasis on computationally demanding high-dimensional cases. Finally, we report efficiency results and ablation studies to assess the contribution of each module.


\subsection{Experiment Setup}
We evaluate K$^2$SVD against state-of-the-art (SOTA) baselines on real-world and dynamic forecasting tasks. To make the comparison clear, we organize the datasets by their dimensionality.

\textbf{Datasets.} We conduct experiments on nine forecasting datasets: eight real-world datasets from ProbTS \cite{zhang2024probts}, a comprehensive benchmark for probabilistic forecasting; and one dynamical-system dataset following the experimental setting of \cite{jeong2025efficient}. The high-dimensional group consists of Electricity-L, Weather-L, and Ordered-MNIST. Electricity-L contains the consumption of electricity in kWh, Weather-L contains multivariate climatological measurements, and Ordered-MNIST represents image-based dynamic observations. With tens or hundreds of dimensions, these datasets better assess whether the learned low-rank Koopman space improves scalability and efficiency. The low-dimensional group includes ETTh1-L, ETTh2-L, ETTm1-L, ETTm2-L, Exchange-L, and ILI-L. These datasets contain relatively few variables and evaluate whether K$^2$SVD remains competitive when the observation space is already compact. Detailed dataset statistics are reported in Table~\ref{tab:stat}.

The context and horizon length for the main results are displayed in table~\ref{tab:stat} and additional results for longer term time series prediction are shown in the Appendix~\ref{sec:lt}. 

\begin{table*}[h]
\centering
\caption{Statistics and Description of the nine datasets.}
\resizebox{\textwidth}{!}{%
\begin{tabular}{l|l|rccccrl}
\toprule
\textbf{Dimension Group} & \textbf{Dataset} & \textbf{\#var.} & \textbf{range} & \textbf{freq.} & \textbf{Context} & \textbf{Horizon} & \textbf{Description} \\
\midrule
\multirow{3}{*}{\textbf{High-dimensional}}
 & Electricity-L & 321 & $\mathbb{R}^+$ & H        & 96 & 96 & Electricity consumption (Kwh) \\
 & Ordered-MNIST  & 28*28   & $(0, 255)$ & NA          & 48 & 48 & Periodic change of handwritten figures \\
 & Weather-L   & 21  & $\mathbb{R}^+$ & 10min      & 96 & 96 & Local climatological data \\
\midrule
\multirow{4}{*}{\textbf{Low-dimensional}}
 & ETTh1/h2-L  & 7   & $\mathbb{R}^+$ & H          & 96 & 96 & Electricity transformer temperature per hour \\
 & ETTm1/m2-L  & 7   & $\mathbb{R}^+$ & 15min      & 96 & 96 & Electricity transformer temperature every 15 min \\
 & Exchange-L  & 8   & $\mathbb{R}^+$ & Busi. Day  & 96 & 96 & Daily exchange rates of 8 countries \\
 & ILI-L  & 7   & $(0, 1)$ & W & 36 & 24 & Ratio of patients seen with influenza-like illness \\

\bottomrule
\end{tabular}%
}
\label{tab:stat}
\end{table*}

\textbf{Baselines.} We compare K$^2$SVD with seven baselines designed for both real-world and dynamic system data. These include four point-forecasting models, $K^2$VAE~\cite{wu2025k2vaekoopmankalmanenhancedvariational}, FITS~\cite{xu2024fits}, iTransformer~\cite{liu2024itransformer}, and Koopa~\cite{liu2023koopa}, as well as three generative models, TSDiff~\cite{kollovieh2023predict}, GRU NVP~\cite{rasul2021multivariate}, and CSDI~\cite{tashiro2021csdi}, with impleameantion details provided in Appendix~\ref{sec:details}.

\textbf{Evaluation Metrics.} We evaluate probabilistic forecasts using NRMSE (Normalized Root Mean Square Error) as defined below. We also report model computational complexity and inference time in the efficiency section to highlight the benefits of the LoRA structure recovered by the proposed K$^2$SVD algorithm.
\begin{equation*}
\mathrm{NRMSE} = \frac{\sqrt{\mathrm{MSE}}} {\frac{1}{N}\sum_{i=1}^{N}|Y_i|},\ \text{with} \  \mathrm{MSE} = \frac{1}{N}\sum_{i=1}^{N} \left(Y_i-\hat{Y}_i\right)^2\!. \label{eq:metric}
\end{equation*}

\subsection{Main Results}\label{subsec:results}
\begin{figure*}[t]
    \centering
    \includegraphics[width=0.9\textwidth]{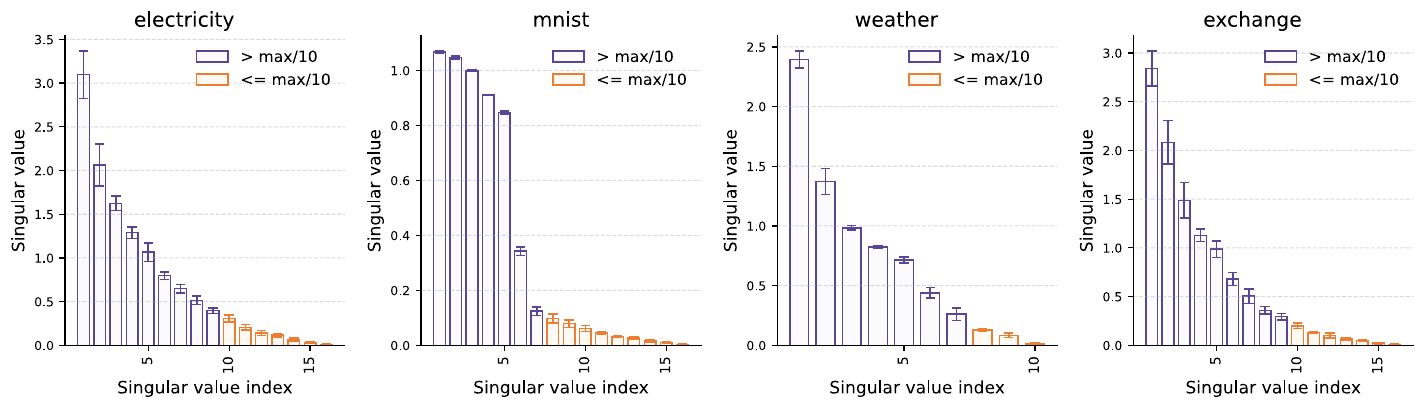}
    \caption{The singular values of the Koopman space exhibit low-rank properties on multiple datasets.}
    \label{fig:spec}
\end{figure*}
\textbf{Empirical evidence for the LoRA property.} We observe low-rank structure across both low-dimensional and high-dimensional datasets, a property that most existing models do not explicitly capture or exploit. As shown by the Koopman-space spectrum in Figure~\ref{fig:spec}, the singular values decay toward zero as the rank increases across all scenarios. Most datasets exhibit this decay within roughly 10-30 dimensions. These observations motivate an adaptive choice of latent dynamic dimension, where we use dynamic dimension 16 for most datasets instead of the commonly chosen 128 in baselines. The ablation study further shows that increasing the latent dimension has only a minor effect on precision but substantially increases computational cost.
One exception is ILI-L, which requires approximately 60 Koopman dimensions before the singular values vanish, indicating a relatively high-rank structure despite its low observation dimension. Results for ILI-L are reported in the Appendix~\ref{sec:hr}.

\begin{table*}[t]
\centering
\caption{Results of models using the NRMSE metric on the high-dimensional datasets and Exchange-L. Best results are shown in \textbf{bold}, and second-best results are \underline{underlined}. '-' indicates that the run took too long to finish.}
\resizebox{0.6\textwidth}{!}{%
\begin{tabular}{c|cccc}
\hline
Model & Electricity-L & MNIST & Weather-L & Exchange-L \\
\hline
FITS
  & 1.3386 ± 0.0038 & 2.0438 ± 0.0007 & 0.6226 ± 0.0133 & 0.0912 ± 0.0001\\
\hline
iTransformer
  & 1.3035 ± 0.0033 & 2.0486 ± 0.0014 & 0.3617 ± 0.0006 & 0.1050 ± 0.0001\\
\hline
Koopa
  & 1.3977 ± 0.0183 & 2.0510 ± 0.0007 & 0.3656 ± 0.0004 & 0.1051 ± 0.0002\\
\hline
TSDiff
  & 1.2107 ± 0.0587 & - & \underline{0.3101 ± 0.0082} & 0.1196 ± 0.0060\\
\hline
GRU NVP
  & 1.0617 ± 0.0299 & - & 0.7841 ± 0.1476 & 0.0662 ± 0.0002\\
\hline
$K^2$VAE
  & \underline{0.8682 ± 0.0106} & \underline{1.9817 ± 0.0379} & \textbf{0.2853 ± 0.0042} & \underline{0.0470 ± 0.0009}\\
\hline
K$^2$SVD
  & \textbf{0.8502 ± 0.0014} & \textbf{1.8125 ± 0.0010} & 0.3124 ± 0.0011 & \textbf{0.0419 ± 0.0004}\\
\hline
\end{tabular}%
}
\label{tab:main_elec}
\end{table*}

\textbf{Prediction performance.}
The prediction results in Table~\ref{tab:main_elec} show that K$^2$SVD performs strongly on Electricity-L, Weather-L, Ordered-MNIST, and Exchange-L. For fairness, a CNN encoder will replace all the encoders when models are applied on MNIST dataset. As described in the dataset setup and summarized in Table~\ref{tab:stat}, the former three datasets form the high-dimensional group, with observations consisting of many interacting variables or image-like dynamic states. K$^2$SVD is well-suited to these settings because it directly identifies a compact transition space from the underlying dynamics, reducing the redundancy and instability of less principled latent embeddings while mitigating the effect of noise through Kalman inference.

For the low-dimensional group, K$^2$SVD also remains competitive, with full results reported in Appendix~\ref{sec:lt}. Exchange-L is included because it is a low-dimensional but challenging financial dataset characterized by weak periodicity and nonstationary dynamics. Together, our results show that the principled Koopman transition space is beneficial across different observation scales, with the efficiency advantage to be evaluated in the next section.


\begin{table*}[t]
\centering
\caption{Comparison of inference time in seconds, and GPU usage in GBs on the high-dimensional datasets. For each metric, the best results are shown in \textbf{bold}, and the second-best results are \underline{underlined}. ``-'' indicates that the run took too long to finish. 
}
\resizebox{0.75\textwidth}{!}{%
\begin{tabular}{c|ccc|ccc}
\hline
\multirow{2}{*}{Model}
& \multicolumn{3}{c|}{Inference Time}
& \multicolumn{3}{c}{GPU Usage}\\
& Electricity-L & MNIST & Weather-L
& Electricity-L & MNIST & Weather-L\\
\hline
FITS
  & 31.261 $\pm$ 1.083
  & 43.497 $\pm$ 0.199
  & 4.072 $\pm$ 0.026
  &  \underline{1.619}
  &  {4.313}
  &  {0.140}\\
\hline
iTransformer
  & 30.674 $\pm$ 0.421
  & 43.276 $\pm$ 0.115
  & 4.177 $\pm$ 0.032
  &  {1.622}
  &  {4.314}
  &  {0.150}\\
\hline
Koopa
  & \underline{29.559 $\pm$ 0.067}
  & \underline{43.097 $\pm$ 0.229}
  & \underline{4.026 $\pm$ 0.020}
  &  { 1.635 }
  &  {4.432}
  &  {0.149}\\
\hline
TSDiff
  & 93.217 $\pm$ 0.572
  & -
  & 13.677 $\pm$ 0.102
  &  {5.692}
  &  - 
  &  {2.276}\\
\hline
GRU NVP
  & 42.229 $\pm$ 0.164
  & -
  & 4.802 $\pm$ 0.069
  &  { 1.817}
  &  -
  &  \underline{0.126}\\
\hline
$K^2$VAE
  & 84.202 $\pm$ 0.167
  & 127.390 $\pm$ 0.255
  & 8.150 $\pm$ 0.126
  &  {1.702}
  &  {4.437}
  &  {0.160}\\
\hline
K$^2$SVD
  & \textbf{10.144 $\pm$ 0.069}
  & \textbf{13.662 $\pm$ 0.205}
  & \textbf{2.787 $\pm$ 0.041}
  &  {\textbf{0.438}}
  &  {\textbf{0.954}}
  &  {\textbf{0.048}}\\
\hline

\end{tabular}%
}
\label{tab:eff}
\end{table*}

\textbf{Efficiency and compactness.} 
From an efficiency perspective, K$^2$SVD establishes SOTA performance in both inference time and GPU memory usage among the evaluated methods, with its advantage becoming particularly pronounced on high-dimensional datasets. All models are evaluated on the same GPU under an identical software environment, and the results are reported in Table~\ref{tab:eff}. K$^2$SVD achieves the fastest inference time on all three datasets. On the two most computationally demanding benchmarks, Electricity and MNIST, it is approximately three times faster than every competing baseline. In terms of memory usage, K$^2$SVD requires less than one-third of the peak GPU memory allocated by the most efficient baseline models. Notably, K$^2$SVD substantially outperforms FITS in both inference speed and memory usage, despite FITS being designed as a low-cost frequency-domain model. This comparison demonstrates that the efficiency of K$^2$SVD arises from its low-rank transition space and from concentrating iterative prediction within this space. Consequently, K$^2$SVD achieves both superior predictive accuracy and substantially lower inference cost, yielding a more favorable accuracy-efficiency trade-off.

\subsection{Ablation Study}
\subsubsection{Ablation on the Dimension of Dynamic Space} \label{subsec:abl1}

We conduct an ablation study to examine the sufficiency of the latent Koopman dimension. Specifically, we vary the rank of the learned Koopman subspace from 16 to 64 to investigate whether reducing the rank harms performance. As shown in Table~\ref{tab:rank_ablation}, performance in prediction precision does not improve significantly yet sometimes drops as the dimension increases, while the total computation measured as FLOPs (floating-point operations per second) booms. Although inference time does not visibly increase, we infer that it is limited by memory bandwidth rather than compute capacity, as the models require no more than 1 GFLOP while the GPU provides over 100 TFLOPS of compute. These results suggest that the dominant system dynamics can be effectively captured in a compact, low-rank Koopman subspace, making higher-dimensional representations redundant.

\begin{table}[h]
\centering
\caption{Ablation study on the Koopman latent dimension. Low-dimensional latent spaces achieve competitive accuracy, indicating redundancy in higher-rank representations.}
\label{tab:rank_ablation}
\resizebox{0.45\textwidth}{!}{
\begin{tabular}{llccc}
\toprule
Dataset & Metric & Rank 16 & Rank 32 & Rank 64 \\
\midrule
\multirow{3}{*}{Exchange-L}
& NRMSE & $0.0441 \pm 0.0016$ & $0.0449 \pm 0.0002$ & $0.0428 \pm 0.0001$ \\
& Inference time (s) & $0.656 \pm 0.027$ & $0.625 \pm 0.044$ & $0.629 \pm 0.036$ \\
& FLOPs & 3M & 6M & 33M \\
\midrule
\multirow{3}{*}{Electricity-L}
& NRMSE & $0.8502 \pm 0.0014$ & $0.906\pm 0.008$ & $0.8690 \pm 0.0120$ \\
& Inference time (s) & $9.786 \pm 0.088$ & $9.980 \pm 0.181$ & $9.856 \pm 0.031$ \\
& FLOPs & 40M & 58M & 137M \\
\bottomrule
\end{tabular}
}
\end{table}

\subsubsection{Ablation on LoRA Objective}
We argue that optimizing the encoder with the LoRA objective provides an efficient and expressive Koopman space. In comparison to end-to-end methods, the encoder trained with the HS objective and frozen in later steps creates a more solid and precise LoRA space, and as the inference length increases it better resists noise and results in more accurate reconstruction. To support this claim, we set the context $t=96$, horizon $T=720$, remove stage 1 and directly train the encoder, Koopman matrix $\mathsf{K}_{\theta}$, and Kalman parameters in an end-to-end theme with only $\mathcal{L}_e$. As shown in Table~\ref{tab:abl1}, the training theme with LoRA objective achieves better performance under the same hidden-space dimension on the datasets MNIST, Electricity and Exchange, and comparable results under weather.

\begin{table}[h]
\centering
\caption{Ablation study on LoRA objective shows that the first training stage has advantages in prediction performance.}
\resizebox{0.45\textwidth}{!}{%
\begin{tabular}{c|cccc}
\hline
Ablation  & Electricity-L & MNIST & Weather-L & Exchange-L \\
\hline
End-to-end
  & 1.6286 $\pm$ 0.1900
  & 1.9759 ± 0.0718
  & \textbf{0.3380 ± 0.0013} 
  & 0.1239 ± 0.0019 
   \\
\hline
2-Stage
  & \textbf{1.3035 ± 0.0155} 
  & \textbf{1.9423 ± 0.0029}
  & 0.3470 ± 0.0009 
  
  & \textbf{0.1139 ± 0.0010} 
  \\
\hline
\end{tabular}%
}
\label{tab:abl1}
\end{table}

\subsubsection{Ablation on Kalman inference}\label{sec:abl_kalman}
As we argued in section~\ref{sec:Kalman}, the output of the Koopman module, the linear rollout of the base vector in prediction spaces, contains noise from various sources, and we applied Kalman inference to handle it. To justify the necessity, we use the linear rollout as the predicted base vectors and directly decode it to get the prediction in data space. The results show different degrees of decline in performance, which reveals the noise in linear prediction and emphasizes the importance of Kalman inference. 

\begin{table}[h]
\centering
\caption{Ablation study on Kalman inference shows the noisy property of linear prediction and the advantage of filtering.}
\resizebox{0.45\textwidth}{!}{
\begin{tabular}{c|cccc}
\hline
Ablation  & Electricity-L  & MNIST & Weather-L & Exchange-L  \\
\hline
Linear
  & 3.0902 $\pm$ 0.3347
  & 1.8829 ± 0.0054
  & 0.4228 $\pm$ 0.0119
  & 0.0447 $\pm$ 0.0005
   \\
\hline
Filtered
  & \textbf{0.8502 ± 0.0014}
  & \textbf{1.8125 ± 0.0010}
  & \textbf{0.3124 ± 0.0011} 
  & \textbf{0.0419 $\pm$ 0.0004}
  \\
\hline
\end{tabular}
}
\label{tab:abl2}
\end{table}

\section{Conclusion}
This work revisits time-series forecasting from the perspective of operator structure rather than only predictive capacity. The central observation is that many forecasting benchmarks contain a compact transition structure that can be exposed through the spectrum of the learned Koopman space. By building the forecasting model around this structure, K$^2$SVD reduces the latent dynamic dimension substantially while retaining competitive prediction accuracy and achieving SOTA inference efficiency, strongly exceeds baseline models. 

The experiments also show results with clear distinction in intrinsically different datasets, revealing information about the data and provides visible judge through the learned spectrum rather than empirical observations.

More broadly, these results suggest that efficient forecasting should not only compress model parameters, but also identify which part of the temporal dynamics is worth modeling explicitly. Future work can extend this direction by allowing the Koopman rank and transition operator to adapt across time, and by developing uncertainty mechanisms that separate random perturbations from genuine changes in the dynamics.

\clearpage

\bibliography{ref}      
\bibliographystyle{plainnat}   

\clearpage
\onecolumn
\appendix
\section{Proof to the LoRA Loss}\label{sec:proof}

Here, we provide the proof to justify the LoRA loss proposed in \cite{jeong2025efficient}.

Recall that $\mathcal{L}_{\text{lora}}(\mathbf{f}, \mathbf{g}) \triangleq -2\,\text{tr}(\text{T}[\mathbf{f}, \mathbf{g}]) + \text{tr}(\text{M}_{\rho_0}[\mathbf{f}]\text{M}_{\rho_1}[\mathbf{g}])$. To prove, first note that

\begin{align}
     \left\| \mathcal{K} - \sum_{i=1}^{k} f_i \otimes g_i \right\|_{\text{HS}}^2 - \|\mathcal{K}\|_{\text{HS}}^2= -2\sum_{i=1}^{k} \langle f_i, \mathcal{K} g_i \rangle_{\rho_0}
    + \sum_{i=1}^{k}\sum_{j=1}^{k} \langle f_i, f_j \rangle_{\rho_0} \langle g_i, g_j \rangle_{\rho_1}.
\end{align}

Here, the first term can be rewritten as
\begin{align}
     \sum_{i=1}^{k} \langle f_i, \mathcal{K} g_i \rangle_{\rho_0}
&= \sum_{i=1}^{k} \mathbb{E}_{\rho_0(\mathbf{x})p(\mathbf{x}'|\mathbf{x})} \bigl[ f_i(\mathbf{x}) g_i(\mathbf{x}') \bigr]\\
&= \text{tr}\!\left( \mathbb{E}_{\rho_0(\mathbf{x})p(\mathbf{x}'|\mathbf{x})} \bigl[ \mathbf{f}(\mathbf{x})\mathbf{g}(\mathbf{x}')^\top \bigr] \right)
= \text{tr}(\text{T}[\mathbf{f}, \mathbf{g}]).
\end{align}

Likewise, the second term is
\begin{align}
    \sum_{i=1}^{k}\sum_{j=1}^{k} \langle f_i, f_j \rangle_{\rho_0} \langle g_i, g_j \rangle_{\rho_1}
&= \text{tr}\!\left( \mathbb{E}_{\rho_0(\mathbf{x})} \bigl[ \mathbf{f}(\mathbf{x})\mathbf{f}(\mathbf{x})^\top \bigr]
  \mathbb{E}_{\rho_1(\mathbf{x}')} \bigl[ \mathbf{g}(\mathbf{x}')\mathbf{g}(\mathbf{x}')^\top \bigr] \right)\\
&= \text{tr}(\text{M}_{\rho_0}[\mathbf{f}]\text{M}_{\rho_1}[\mathbf{g}]).
\end{align}

This concludes the proof.

\section{Algorithm}\label{ref:algo}
\begin{algorithm*}[t]
\caption{Two-Stage K$^2$SVD Training and Forecasting}
\label{alg:k2svd}
\KwIn{Training pairs $(X,Y)$; patch length $s$; Koopman rank $d$; variant $\nu\in\{\text{static},\text{dynamic}\}$}
\KwOut{Encoders $\mathbf{f},\mathbf{g}$; Koopman matrix $\mathsf{K}$; Kalman parameters $A,H,Q,R$; decoder coefficients $\mathbf{z}_{\mathbf{b}}^{\mathrm{ols}}$}

\BlankLine
\textbf{Stage 1: Learn a low-rank Koopman space}\;
\For{each training batch $X$}{
    Split $X$ into non-overlapping patches $X^P=[\mathbf{x}_1^P,\ldots,\mathbf{x}_n^P]$\;
    Compute shifted Koopman features
    $\tilde{\mathbf{f}}=[\mathbf{f}(\mathbf{x}_2^P),\ldots,\mathbf{f}(\mathbf{x}_n^P)]$ and
    $\tilde{\mathbf{g}}=[\mathbf{g}(\mathbf{x}_1^P),\ldots,\mathbf{g}(\mathbf{x}_{n-1}^P)]$\;
    Evaluate
    $\mathcal{L}_{\mathrm{LoRA}}
    \leftarrow
    -2\,\mathrm{tr}(\mathsf{T}[\tilde{\mathbf{f}},\tilde{\mathbf{g}}])
    + \mathrm{tr}(\mathsf{M}_{\rho_0}[\tilde{\mathbf{f}}]\mathsf{M}_{\rho_1}[\tilde{\mathbf{g}}])$\;
    Update $\mathbf{f},\mathbf{g}$ by minimizing $\mathcal{L}_{\mathrm{LoRA}}$\;
}

\BlankLine
\textbf{Initialize finite-dimensional dynamics}\;
Choose the learned basis $\mathbf{b}\in\{\mathbf{f},\mathbf{g}\}$ used for forecasting\;
Estimate the regression-induced Koopman matrix
$\mathsf{K}_{\mathbf{b}}^{\mathrm{ols}}
\leftarrow
\bigl(\hat{\mathsf{M}}_{\rho_0}[\mathbf{b}]\bigr)^{\dagger}
\hat{\mathsf{T}}[\mathbf{b}]$
using training batches\;
Initialize $A\leftarrow\mathsf{K}_{\mathbf{b}}^{\mathrm{ols}}$, $H\leftarrow I$, and noise covariances $Q,R$\;
\eIf{$\nu=\text{dynamic}$}{
    Initialize trainable $\mathsf{K}_{\theta}\leftarrow\mathsf{K}_{\mathbf{b}}^{\mathrm{ols}}$ and set $\mathsf{K}\leftarrow\mathsf{K}_{\theta}$\;
}{
    Set $\mathsf{K}\leftarrow\mathsf{K}_{\mathbf{b}}^{\mathrm{ols}}$ and keep $\mathsf{K}$ fixed\;
}
Compute decoder coefficients
$\mathbf{z}_{\mathbf{b}}^{\mathrm{ols}}
\leftarrow
(\mathsf{M}_{\rho_0}[\mathbf{b}])^{\dagger}\langle h,\mathbf{b}\rangle_{\rho_0}$ with $h(\mathbf{x})=\mathbf{x}$\;

\BlankLine
\textbf{Stage 2: Train Kalman forecasting module}\;
Freeze $\mathbf{f},\mathbf{g}$\;
\For{each training batch $(X,Y)$}{
    Encode the final context patch to obtain the initial latent state $\hat{\mathbf{b}}_{n|n}$\;
    Roll out the regression-induced observation sequence $\mathbf{b}^{L}$ using $\mathsf{K}_{\mathbf{b}}^{\mathrm{ols}}$\;
    \For{each forecast patch index $k=n+1,\ldots,n+m$}{
        Predict latent state and covariance:
        $\hat{\mathbf{b}}_{k|k-1}\leftarrow A\hat{\mathbf{b}}_{k-1|k-1}$,
        $\Sigma_{k|k-1}\leftarrow A\Sigma_{k-1|k-1}A^\top+Q$\;
        Set observation $\mathbf{o}_k\leftarrow\mathbf{b}^{L}[k]$ and compute Kalman gain
        $G_k\leftarrow\Sigma_{k|k-1}H^\top(H\Sigma_{k|k-1}H^\top+R)^{-1}$\;
        Update latent state and covariance:
        $\hat{\mathbf{b}}_{k|k}\leftarrow\hat{\mathbf{b}}_{k|k-1}+G_k(\mathbf{o}_k-H\hat{\mathbf{b}}_{k|k-1})$,
        $\Sigma_{k|k}\leftarrow(I-G_kH)\Sigma_{k|k-1}$\;
    }
    Decode patch forecasts $\hat{X}^{P}\leftarrow\hat{\mathbf{b}}^\top\mathbf{z}_{\mathbf{b}}^{\mathrm{ols}}$ and de-patch to obtain $\hat{X}$\;
    Compute
    $\mathcal{L}_{e}
    \leftarrow
    \|Y-\hat{X}\|_2^2
    +D_{KL}\!\left(\mathbb{Q}(\hat{\mathbf{b}}\mid X)\,\|\,\mathbb{P}(\mathbf{b}\mid X)\right)$\;
    \eIf{$\nu=\text{dynamic}$}{
        Jointly update $\mathsf{K}_{\theta}$ and $A,H,Q,R$ using $\mathcal{L}_{e}$\;
    }{
        Update $A,H,Q,R$ by minimizing $\mathcal{L}_{e}$\;
    }
}
\end{algorithm*}

Algorithm~\ref{alg:k2svd} summarizes the two-stage training procedure described in the method section. The first stage learns an operator-consistent low-rank Koopman space using the LoRA objective. The second stage freezes this learned Koopman basis and trains the Kalman forecasting module. The static variant K$^2$SVD-S keeps the regression-induced Koopman matrix fixed, while K$^2$SVD further updates the finite-dimensional Koopman matrix during downstream forecasting training.

\section{Experimental Settings}
\label{sec:details}

All experiments are conducted on a single GPU with a global random seed of (1, 2, 42, 123, 2026) to ensure reproducibility and for constructing error bar.
The model is trained for up to 50 epochs, with each epoch consisting of 100 training batches. Detailed setting to bu found in code. 
We use a gradient clipping threshold of 0.5 to ensure training stability and set the learning rate to $1\times10^{-3}$.
Model checkpoints and logs are saved to a local results directory, and validation is performed at the end of every epoch.

\paragraph{Two-Stage Training.}
We employ a two-stage training curriculum managed by a dedicated callback.
The first stage spans 15 epochs, during which the model undergoes a warm-up phase to effectively initialize the Koopman operator.
The remaining 35 epochs constitute the second stage, in which the full model is optimized end-to-end.

\paragraph{Model Configuration.}
The data preprocessing follows the K$^2$VAE architecture~\cite{wu2025k2vaekoopmankalmanenhancedvariational}, as input sequence is divided into non-overlapping patches of length 24.
The Koopman latent space has a dynamic dimension of 16, initialized using a dynamic Koopman scheme, with multistep prediction enabled.
The encoder and decoder each consist of 3 hidden layers of width 256.
The VAE regularization weight is set to $\beta = 0.01$, imposing a light KL penalty to preserve the expressiveness of the latent space.
During inference, the model draws 100 samples per input to construct the predictive distribution, which is summarized over 20 quantiles.
The sampling schedule runs for 20 steps.

\paragraph{Data Pipeline.}
Training and evaluation use batch sizes of 64 and 32, respectively, with 8 parallel data-loading workers.
All input features are normalized using standard scaling (zero mean, unit variance) prior to model input.
A held-out validation split is partitioned from the training set to monitor generalization during training.

\section{Additional Empirical Results}\label{sec:hd}
\subsection{Additional Results on Long-Term Time Series Prediction}\label{sec:lt}

Tables~\ref{tab:low_dim_results} and~\ref{tab:high_dim_results} summarize the NRMSE results for prediction lengths 96 and 720. Table~\ref{tab:low_dim_results} reports the low-dimensional datasets, including Exchange-L and the ETT series. These datasets have compact observation spaces, so the main question is whether the learned Koopman transition remains stable when the prediction horizon increases. K$^2$SVD performs competitively across this group and obtains the best results on Exchange-L for both prediction lengths, showing that the learned operator-induced transition space is useful even when the raw data dimension is small. On the ETT datasets, K$^2$SVD is strongest on ETTh1-L at prediction length 96 and on ETTh1-L, ETTh2-L, and ETTm2-L at prediction length 720, while $K^2$VAE and Koopa remain strong competitors on several shorter-horizon or lower-variance cases.

Table~\ref{tab:high_dim_results} reports the high-dimensional datasets, including Electricity-L, Ordered-MNIST, and Weather-L. These datasets are the most relevant for evaluating the computational motivation of K$^2$SVD, because the observations contain many coupled variables or image-like dynamic states. K$^2$SVD achieves the best results on Electricity-L and Ordered-MNIST at prediction length 96 and remains competitive at prediction length 720. On Weather-L, $K^2$VAE attains the best NRMSE, while K$^2$SVD remains close to the strongest baselines. Together with the efficiency results in the main text, these results suggest that K$^2$SVD provides a favorable accuracy-efficiency trade-off: it is not uniformly best on every dataset, but it is especially effective when a compact Koopman transition captures the dominant dynamics.

\subsection{Additional Results on Long Prediction Efficiency}
We further evaluate long-horizon inference efficiency to examine whether the computational advantage of K$^2$SVD persists when prediction requires many recursive rollout steps. The comparison includes FITS, iTransformer, Koopa, and $K^2$VAE because they represent the main classes of efficiency-competitive baselines used in this work. This set therefore compares K$^2$SVD against both general-purpose long-sequence predictors and methods that are structurally closest to it.

Table~\ref{tab:long_eff} shows that the advantage of K$^2$SVD becomes more pronounced under long prediction lengths. K$^2$SVD achieves the fastest inference on all three datasets, reducing inference time by about $2.4\times$--$5.5\times$ compared with the strongest non-K$^2$SVD baseline. Its GPU usage is also consistently the lowest, especially on MNIST, where the compact Koopman transition avoids carrying the high-dimensional observation space through each rollout step. The FLOPs comparison further highlights the source of this efficiency: after the encoder maps observations into the learned low-rank Koopman space, long-range prediction is performed by lightweight linear state evolution and Kalman inference rather than repeated high-dimensional nonlinear transformations. These results support the central design of K$^2$SVD: by explicitly learning a compact operator-induced transition space, the model improves not only predictive accuracy but also the time, memory, and arithmetic cost of long-horizon forecasting.

\begin{table*}[t]
\centering
\caption{Comparison of inference time in seconds, GPU usage in GBs, and quantification of computation by meta FLOPs on the selected high and low dimensional datasets when inference length is $T=480$ for MNIST and $T=720$ for the rest. For each metric, the best results are shown in \textbf{bold}, and the second-best results are \underline{underlined}. ``-'' indicates that the run took too long to finish. 
}
\resizebox{\textwidth}{!}{%
\begin{tabular}{c|ccc|ccc|ccc}
\hline
\multirow{2}{*}{Model}
& \multicolumn{3}{c|}{Inference Time}
& \multicolumn{3}{c}{GPU Usage}
& \multicolumn{3}{c}{FLOPs} \\
& Electricity-L & MNIST & Weather-L
& Electricity-L & MNIST & Weather-L
& Electricity-L & MNIST & Weather-L \\
\hline
FITS
  & 168.5 ± 4.654
  & 564.423 ± 12.195
  & 17.968 ± 4.434
  &   8.723
  &   24.407
  &  0.938
  &  42881.1
  &  104731.4
  &  2805.3 \\
\hline
iTransformer
  &  143.538 ± 11.071
  &  380.082 ± 0.901
  &  15.104 ± 4.055
  &  {8.724 }
  &  {24.408}
  &  0.951
  &  47211.8
  &  121302.3
  &  2411.5 \\
\hline
Koopa
  & 131.898 ± 11.382
  & 453.949 ± 21.723
  & 15.010 ± 3.968
  &  { 8.739 }
  &  {24.433}
  &  0.948
  &  {31551.6}
  &  {76572.8}
  &  2380.1 \\
\hline
$K^2$VAE
  & 488.162 ± 8.377
  & 2228.906 ± 74.189
  & 45.667 ± 6.598
  &  {8.810}
  &  {24.477}
  &  {0.965}
  &  {108428.2}
  &  {75946.1}
  &  {92328.9} \\
\hline
K$^2$SVD
  & \textbf{44.999 ± 0.063}
  & \textbf{120.644 ± 0.014}
  & \textbf{6.184 ± 0.039}
  &  {\textbf{1.910}}
  &  {\textbf{5.052}}
  &  {\textbf{0.210}}
  &  \textbf{19899.0}
  &  \textbf{9279.3}
  &  \textbf{282.2} \\
\hline

\end{tabular}%
}
\label{tab:long_eff}
\end{table*}

\begin{table*}[t]
\centering
\caption{Results of models using the NRMSE metric on low-dimensional datasets with prediction lengths 96 and 720. Best results are shown in \textbf{bold}, and second-best results are \underline{underlined}. '*' indicates that the model encounters invalid computation due to instability.}
\resizebox{\textwidth}{!}{%
\begin{tabular}{c|cc|cc|cc|cc|cc}
\hline
\multirow{2}{*}{Model}
& \multicolumn{2}{c|}{Exchange-L}
& \multicolumn{2}{c|}{ETTh1-L}
& \multicolumn{2}{c|}{ETTh2-L}
& \multicolumn{2}{c|}{ETTm1-L}
& \multicolumn{2}{c}{ETTm2-L} \\
& 96 & 720 & 96 & 720 & 96 & 720 & 96 & 720 & 96 & 720 \\
\hline
FITS
  & 0.0912 ± 0.0001 & \underline{0.1233 $\pm$ 0.0030}
  & 0.6673 ± 0.0026 & 0.7967 $\pm$ 0.0160
  & 0.3349 ± 0.0006 & 0.6068 $\pm$ 0.0277
  & 0.5985 ± 0.0016 & \underline{0.6674 $\pm$ 0.0087}
  & 0.2937 ± 0.0027 & 0.9151 ± 0.0695\\
\hline
iTransformer
  & 0.1050 ± 0.0001 & 0.1581 $\pm$ 0.0020
  & 0.6592 ± 0.0011 & 0.7152 $\pm$ 0.0010
  & 0.3397 ± 0.0015 & 2.9661 $\pm$ 0.9167
  & 0.6643 ± 0.0071 & 0.8523 $\pm$ 0.0278
  & 0.3132 ± 0.0036 & 0.7747 ± 0.0898\\
\hline
Koopa
  & 0.1051 ± 0.0002 & 0.1831 $\pm$ 0.0029
  & 0.6832 ± 0.0040 & \underline{0.7117 $\pm$ 0.0045}
  & 0.3568 ± 0.0010 & \underline{0.3897 $\pm$ 0.0049}
  & 0.6204 ± 0.0066 & \textbf{0.6376 ± 0.0016}
  & * & \underline{0.3572 ± 0.0052}\\
\hline
TSDiff
  & 0.1196 ± 0.0060 & 0.1632 $\pm$ 0.0087
  & 0.8040 ± 0.0059 & 1.0635 $\pm$ 0.0437
  & 0.5095 ± 0.0191 & 0.6274 $\pm$ 0.0022
  & 0.6668 ± 0.0172 & 0.9958 $\pm$ 0.0154
  & 0.4048 ± 0.0245 & 0.5154 ± 0.0097\\
\hline
GRU NVP
  & 0.0662 ± 0.0002 & 0.1935 $\pm$ 0.0097
  & 0.8307 ± 0.0249 & 0.9291 $\pm$ 0.0509
  & 0.6020 ± 0.0639 & 8.1696 $\pm$ 2.1317
  & 1.0492 ± 0.0669 & 3.6860 $\pm$ 0.7297
  & 0.3811 ± 0.0319 & 2.0532 ± 1.0939\\
\hline
CSDI
  & 0.0620 ± 0.0030 & 0.2099 $\pm$ 0.0096
  & 0.8884 ± 0.0495 & 1.0901 $\pm$ 0.0181
  & \underline{0.3205 ± 0.0162} & 0.4870 $\pm$ 0.0143
  & 0.5719 ± 0.0068 & -
  & \underline{0.2162 ± 0.0025} & 0.3873 ± 0.0030\\
\hline
$K^2$VAE
  & \underline{0.0470 ± 0.0009} & \underline{0.1145 $\pm$ 0.0012}
  & \underline{0.6231 ± 0.0062} & 0.7199 $\pm$ 0.0144
  & \textbf{0.2895 ± 0.0035} & 0.4678 $\pm$ 0.0353
  & \textbf{0.5330 ± 0.0054} & 0.6810 $\pm$ 0.0090
  & \textbf{0.2131 ± 0.0006} & 0.3662 ± 0.0125\\
\hline

\hline
K$^2$SVD
  & \textbf{0.0419 ± 0.0004} & \textbf{0.1139 $\pm$ 0.0010}
  & \textbf{0.6124 ± 0.0091} & \textbf{0.7064 ± 0.0012}
  & 0.3251 ± 0.0157 & \textbf{0.3476 ± 0.0006}
  & \underline{0.5555 ± 0.0010} & 0.6973 ± 0.0008
  & 0.2283 ± 0.0011 & \textbf{0.3273 ± 0.0004}\\
\hline
\end{tabular}%
}
\label{tab:low_dim_results}
\end{table*}

\begin{table*}[h]
\centering
\caption{Results of models using the NRMSE metric on high-dimensional datasets with prediction lengths 96 and 720. Best results are shown in \textbf{bold}, and second-best results are \underline{underlined}. '-' indicates that the model takes too long to finish training.}
\resizebox{\textwidth}{!}{%
\begin{tabular}{c|cc|cc|cc}
\hline
\multirow{2}{*}{Model}
& \multicolumn{2}{c|}{Electricity-L}
& \multicolumn{2}{c|}{MNIST}
& \multicolumn{2}{c}{Weather-L} \\
& 96 & 720 & 48 & 360 & 96 & 720 \\
\hline
FITS
& 1.3386 ± 0.0038 & 2.2749 $\pm$ 0.1578
& 2.0438 ± 0.0007 & 2.0486 $\pm$ 0.0004
& 0.6226 ± 0.0133 & 0.7892 $\pm$ 0.0018\\
\hline
iTransformer
& 1.3035 ± 0.0033 & 1.3560 ± 0.0063
& 2.0486 ± 0.0014 & \underline{2.0336 $\pm$ 0.0003}
& 0.3617 ± 0.0006 & 0.3769 $\pm$ 0.0000\\
\hline
Koopa
& 1.3977 ± 0.0183 & 1.3979 $\pm$ 0.0026
& 2.0510 ± 0.0007 & 2.0428 $\pm$ 0.0004
& 0.3656 ± 0.0004 & 0.3912 $\pm$ 0.0043\\
\hline
TSDiff
& 1.2107 ± 0.0587 & -
& - & -
& \underline{0.3101 ± 0.0082} & 0.3623 $\pm$ 0.0237\\
\hline
GRU NVP
& 1.0617 ± 0.0299 & -
& - & -
& 0.7841 ± 0.1476 & 0.6221 $\pm$ 0.1136\\
\hline
$K^2$VAE
& \underline{0.8682 ± 0.0106} & \textbf{1.0890 $\pm$ 0.0084}
& \underline{1.9817 ± 0.0379} & 2.0353 $\pm$ 0.00002
& \textbf{0.2853 ± 0.0042} & \textbf{0.3084 $\pm$ 0.0025}\\
\hline
K$^2$SVD
& \textbf{0.8502 ± 0.0014} & \underline{1.3035 ± 0.0155}
& \textbf{1.8125 ± 0.0010} & \textbf{1.9423 ± 0.0029}
& 0.3124 ± 0.0011 & \underline{0.3470 ± 0.0009}\\
\hline
\end{tabular}%
}
\label{tab:high_dim_results}
\end{table*}

\subsection{Discussion of High-rank Dataset}\label{sec:hr}
Although ILI-L is low-dimensional in its observation space, Figure~\ref{fig:ili} shows that its Koopman spectrum decays more slowly than the other datasets. We therefore report it separately as a high-rank case. Table~\ref{tab:ili_results} shows that K$^2$SVD remains comparable but does not dominate this dataset. This behavior is consistent with the spectral evidence: when the effective Koopman rank is high, a compact latent space necessarily discards more non-negligible singular directions, so the resulting finite-dimensional transition cannot preserve as much of the predictive dynamics as it does on low-rank datasets.

ILI-L also exhibits properties that are less aligned with the assumptions behind the compact Koopman approximation used in K$^2$SVD. Influenza-like illness time series can undergo abrupt changes across epidemic phases, changes in dominant temporal patterns, and distribution shifts between context and forecast windows. Prior Koopman forecasting work has emphasized that temporal distribution shifts and non-stationary dynamics are central challenges in real-world time-series prediction~\cite{wang2022knf,liu2023koopa}. In such cases, a single low-dimensional transition operator is harder to estimate reliably: the regression-induced operator may be biased toward the context-window dynamics, and recursive long-horizon rollout can amplify this mismatch. Methods with stronger local adaptation, generative flexibility, or larger effective latent capacity can therefore be more favorable on ILI-L. We view this result as an informative limitation rather than a contradiction of the main findings: K$^2$SVD is designed to exploit datasets with dominant low-rank Koopman structure, while ILI-L represents a high-rank and distribution-shifted regime where that structural advantage is weaker.

\begin{table*}[h]
\centering
\begin{minipage}[t]{0.4\textwidth}
    \centering
    \vspace{0pt}
    \caption{Results of models using the NRMSE metric on ILI-L with prediction lengths 96 and 720. Best results are shown in \textbf{bold}, and second-best results are \underline{underlined}.}
    \resizebox{\linewidth}{!}{%
    \begin{tabular}{c|cc}
    \hline
    Model & 24 & 60 \\
    \hline
    FITS & 0.7467 ± 0.0288 & 0.7199 $\pm$ 0.0256\\
    \hline
    iTransformer & 0.7834 ± 0.0125 & 0.7791 $\pm$ 0.0221\\
    \hline
    Koopa & 1.0881 ± 0.0822 & 0.9026 $\pm$ 0.0146\\
    \hline
    TSDiff & 0.6929 ± 0.0246 & 0.7332 $\pm$ 0.0068\\
    \hline
    GRU NVP & \textbf{0.2934 ± 0.0269} & \underline{0.4328 $\pm$ 0.0258}\\
    \hline
    CSDI & 0.7214 ± 0.0416 & 0.7856 $\pm$ 0.0211\\
    \hline
    $K^2$VAE & \underline{0.2971 ± 0.0103} & \textbf{0.4246 $\pm$ 0.0051}\\
    \hline
    K$^2$SVD & 0.4579 ± 0.0042 & 0.4630 ± 0.0049\\
    \hline
    \end{tabular}%
    }
    \label{tab:ili_results}
\end{minipage}
\hfill
\begin{minipage}[t]{0.48\textwidth}
    \centering
    \vspace{0pt}
    \includegraphics[width=\linewidth]{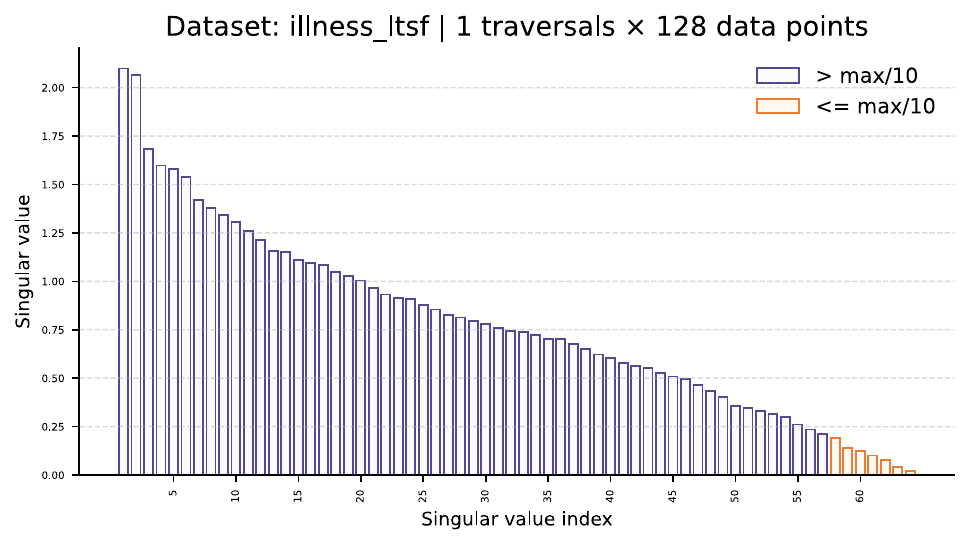}
    \captionof{figure}{The singular values of the Koopman space exhibit higher rank properties on ILI-L.}
    \label{fig:ili}
\end{minipage}

\end{table*}

\subsection{Additional Ablation study}
\subsubsection{Ablation on estimation of Koopman}

As described in the method section, K$^2$SVD applied a trainable Koopman matrix as the LoRA estimation as operator. To show the benefit, we ablate the trainable matrix while initializing and fixing the Koopman matrix as the least-squares estimate $\mathsf{K}_{\mathbf{b}}^{\mathrm{ols}}$ and then optimizes only the Kalman filter parameters. We denote the static matrix theme as K$^2$SVD-S and shows its results on multiple datasets. 
Table~\ref{tab:scheme_ablation} shows that K$^2$SVD consistently improves over K$^2$SVD-S on the evaluated datasets. This improvement is consistent with the motivation in the overall algorithm: the static regression-induced matrix is obtained from finite-sample one-step latent regression, which can be affected by noise and model mismatch.

\begin{table}[h]
\centering
\caption{Ablation study on the two training schemes using the NRMSE metric. K$^2$SVD-S uses the regression-induced Koopman matrix, while K$^2$SVD dynamically updates the Koopman matrix during downstream forecasting training.}
\label{tab:scheme_ablation}
\resizebox{0.6\textwidth}{!}{%
\begin{tabular}{c|cccc}
\hline
Model & Electricity-L & MNIST & Weather-L & Exchange-L \\
\hline
K$^2$SVD-S
  & 1.0417 ± 0.0692 & 2.0134 ± 0.0017 & 0.3209 ± 0.0016 & 0.0441 ± 0.0016\\
\hline
K$^2$SVD
  & \textbf{0.8502 ± 0.0014} & \textbf{1.8125 ± 0.0010} & \textbf{0.3124 ± 0.0011} & \textbf{0.0419 ± 0.0004}\\
\hline
\end{tabular}%
}
\end{table}

\end{document}

%% file: styles.tex
\documentclass[letterpaper]{article}

\PassOptionsToPackage{authoryear,round}{natbib}

\usepackage[preprint]{aaai2027}
\usepackage[hyphens]{url} 
\usepackage{graphicx} 
\usepackage{natbib} 
\setcitestyle{authoryear,round,aysep={ }}
\makeatletter
\newif\ifkTwoSVD@bibstylewritten
\let\kTwoSVD@origbibliographystyle\bibliographystyle
\renewcommand{\bibliographystyle}[1]{%
  \def\kTwoSVD@requestedstyle{#1}%
  \def\kTwoSVD@aaaistyle{aaai2027}%
  \ifx\kTwoSVD@requestedstyle\kTwoSVD@aaaistyle
  \else
    \ifkTwoSVD@bibstylewritten
    \else
      \kTwoSVD@origbibliographystyle{#1}%
      \global\kTwoSVD@bibstylewrittentrue
    \fi
  \fi
}
\AtBeginDocument{%
  \let\kTwoSVD@origbibliography\bibliography
  \renewcommand{\bibliography}[1]{%
    \ifkTwoSVD@bibstylewritten
    \else
      \kTwoSVD@origbibliographystyle{plainnat}%
      \global\kTwoSVD@bibstylewrittentrue
    \fi
    \kTwoSVD@origbibliography{#1}%
  }%
}
\makeatother
\usepackage{caption} 
\usepackage[utf8]{inputenc} 
\usepackage[T1]{fontenc}    
\usepackage{url}            
\usepackage{booktabs}       
\usepackage{amsfonts}       
\usepackage{nicefrac}       
\usepackage{microtype}      
\usepackage{xcolor}         
\usepackage{multirow}
\usepackage{graphicx}
\usepackage{subfigure}
\usepackage{amsmath,amssymb}
\usepackage{enumitem}
\usepackage{bm}
\usepackage{physics}
\usepackage{subcaption}
\usepackage{amssymb}
\usepackage{tikz}
\usepackage{svg}
\usetikzlibrary{arrows.meta, positioning, shapes, fit}
\usepackage[ruled,vlined]{algorithm2e}
\usepackage{amssymb}
\usepackage{framed}

\newcommand{\EQ}{\begin{equation}}
\newcommand{\EN}{\end{equation}}
\newcommand{\ben}{\begin{enumerate}}
\newcommand{\een}{\end{enumerate}}